# Deep Sequence Models for Text Classification Tasks

Saheed Salahudeen Abdullahi
*Department of Computer Science*
*Kaduna State University*
Kaduna, Nigeria
000-0002-1458-4057

Sun Yiming
*Department of Computer*
*Changchun University of Science and Technology*
Changchun, China
sunyiming@cust.edu.cn

Shamsuddeen Hassan Muhammad
*Department of Computer Science*
*University of Porto*
Porto, Portugal
shmuhammad.csc@buk.edu.ng

Abdulrasheed Mustapha
*Department of Computer Science*
*Maharishi University of Management*
Fairfield, United States
amustapha@miu.edu

Ahmad Muhammad Aminu
*Department of Computer Science*
*Kaduna State University*
Kaduna, Nigeria
muhdaminu@kasu.edu.ng

Abdulkadir Abdullahi
*Department of Computer Science*
*Shehu Shagari College of Education*
Sokoto, Nigeria
rababdul2016@gmail.com

Musa Bello
*Department of Computer Science*
*Kaduna State University*
Kaduna, Nigeria
musa.bello@kasu.edu.ng

Saminu Mohammad Aliyu
*Department of Computer Science*
*Bayero University Kano*
Kano, Nigeria
smaliyu.cs@buk.edu.ng

*Abstract*— The exponential growth of data generated on the Internet in the current information age is a driving force for the digital economy. Extraction of information is the major value in an accumulated big data. Big data dependency on statistical analysis and hand-engineered rules machine learning algorithms are overwhelmed with vast complexities inherent in human languages. Natural Language Processing (NLP) is equipping machines to understand these human diverse and complicated languages. Text Classification is an NLP task which automatically identifies patterns based on predefined or undefined labeled sets. Common text classification application includes information retrieval, modeling news topic, theme extraction, sentiment analysis, and spam detection. In texts, some sequences of words depend on the previous or next word sequences to make full meaning; this is a challenging dependency task that requires the machine to be able to store some previous important information to impact future meaning. Sequence models such as RNN, GRU, and LSTM is a breakthrough for tasks with long-range dependencies. As such, we applied these models to Binary and Multi-class classification. Results generated were excellent with most of the models performing within the range of 80% and 94%. However, this result is not exhaustive as we believe there is room for improvement if machines are to compete with humans.

I. INTRODUCTION

Natural language processing is a field that covers computer understanding, manipulation and generation of human language. The recent rapid growth of information on the Internet with the rapid increase in processing power and algorithms have remarkably advanced NLP in tasks such as Machine translation, Named Entity Recognition, Question Answering, Automatic Speech Recognition, Text Classification, and Abstractive summarization [1].

Text classification or Text Categorization or Topic Spotting or Text Tagging; according to [2] is the task of automatically sorting a set of sentences or documents into classes from a predefined set. It is one of the NLP tasks that is leveraging the advancements in deep learning sequence models. The success of text classification has directly supported sectors such as Health, Law, and Businesses in extracting meanings from the huge volume of texts generated.

However, just like other NLP related tasks, text classification is characterized with vast natural language nuances such as morphology, polysemy, syntax and semantics[3]. While machine learning algorithms including Bag-of-Words, Decision Trees, Naïve Bayes and N-gram are over-whelmed with these nuances; deep learning sequence models such as RNN, GRU and LSTM are surfacing with promising results [4][5]. In addition, sequence models implement memory cell that help the neural network to remember relevant information from the past.

While there are several empirical works on Text classification [6], [7], little attention have being directed towards labelled binary and multiclass classification using various sequence models. We adopt the idea of transferring word embedding from previously trained corpus [5] for similarity and context extraction and use sequence neural models to keep track of relevant information [8]. Finally, we fine-tune the output layer based on the type of classification task. Our primary goal in this study is to use neural sequence models and transfer pretrained word embeddings for binary and multiclass text classification. We accomplish this goal with the following specific contributions:

- Examine the various types of sequence models in deep neural networks such as RNN, GRU and LSTM.

- Apply sequence models to Binary classification with IMDB dataset to train and classify users review sentiment

- Analyze BBC news corpus and apply sequence models for theme extraction.
- Co-relate binary and multi-class classification tasks with deep neural sequence models.

## II. SEQUENCE MODELS

Prior to the emergence of sequence models, text classification have enjoyed relative success with continuous bag-of-words (CBOW) [9], Naïve Bayes, N-grams and Tree-based methods. However, they suffered from defects such as data sparsity, dimension explosion and poor generalization ability. Deep learning recently recorded state of the art success in other related domains by leveraging the enormous data available to learn using neural networks. Those tasks include Machine Translation [8] Automatic Speech Recognition [10], [11], Natural Language Understanding [12], Text Summarization [13], Natural Language Generation [14] and Language Models [15].

Sequence models use deep learning approach to address contextless feature extraction on individual words from its dependent surrounding neighbors. Sequence models automatically extract context-sensitive features from raw text [8]. These features when combined provide a deeper insight to the meaning of the text. The sequence models in text-classification include RNN, GRU and LSTM.

### A. Recurrent Neural Network (RNN)

A Recurrent Neural Network (RNN) is a feedforward neural network which is able to handle a variable-length sequence input. Just like humans, the idea behind RNN is to be able to connect previous information to the present task. RNN comprises of many hidden states with activations, the activations at every time $h_t$ depend on residual of previous time $a_{t-1}$ [16].

More formally, given a sequence $\mathcal{X} = (\mathcal{X}_1, \mathcal{X}_2, \dots, \mathcal{X}_t)$, the RNN updates its recurrent hidden states $h_t$ by:

$$h_t = \begin{cases} 0, & t = 0 \\ \varphi(h_{t-1}, x_t), & \text{otherwise} \end{cases} \quad (1)$$

Where φ = is a nonlinear function such as composition of a logistic regression with an affine transformation.

Traditionally, the hidden state $h_t$ is updated as:

$$h_t = g(Wx_t + h_{t-1}) \quad (2)$$

Where g is a bounded function such as logistic regression, or a tanh function.

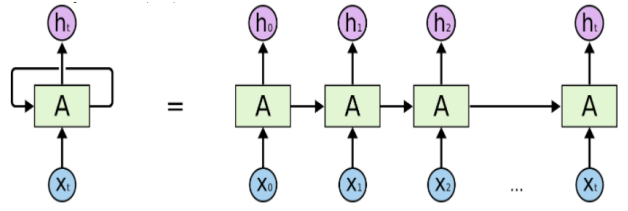

Fig. 1. Recurrent Neural Network (RNN) Structure

However, as neural network tasks continue to evolve, RNN are unable to cope with the task of memorizing previous information. The main reasons were investigated by [17] to be exponential growth and decay during gradient descent.

### B. Long Short Term Memory (LSTM)

The RNN simple strategy used in mapping input sequences to a fixed-sized vector and then feed to a classifier can surfer from exponential decay or growth over long sequences. Hochreiter and Schmidhuber [18] proposed LSTM to specifically address these issues. LSTM is also referred to as *cell state* because it stores information from previous intervals.

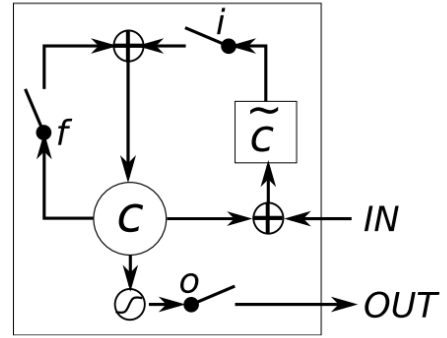

Fig. 2. Long Short Term Memory (LSTM), where c and c' denote the memory cell and new memory cell content, while i, f and o denote the input, forget and output gates, respectively.

Here we provide the general equations of LSTM units at each time step *t* to be a collection of vectors in real number: $i_t$, $f_t$ and $o_t$ accept inputs from $x_t$ and $h_{t-1}$ gates and modulate them with sigmoid [0, 1]. $i_t$ and $f_t$ outputs are symmetrical as they will later be used to decide the output in the cell state $c_t$. $\tilde{c}_t$ takes inputs from $x_t$ and $h_{t-1}$ gates and modulate them with tanh [-1, 1], the output of this gate is a candidate to be stored in $c_t$ based on the output from $i_t$ and $f_t$. New cell state $c_t$ is computed by $i_t \odot \tilde{c}_t$ and $f_t \odot c_t$. Finally $h_t$ from element-wise multiplication of $o_t$ and $c_t$ to expose the internal memory state.

$$i_t = \sigma(W_i x_t + U_i h_{t-1} + V_i c_{t-1}) \quad (3)$$

$$f_t = \sigma(W_f x_t + U_f h_{t-1} + V_f c_{t-1}) \quad (4)$$

$$o_t = \sigma(W_o x_t + U_o h_{t-1} + V_o c_t) \quad (5)$$

$$\tilde{c}_t = tanh(W_c x_t + U_c h_{t-1}) \quad (6)$$

$$c_t = f_t^i \odot c_{t-1} + i_t \odot \tilde{c}_t \quad (7)$$
$$h_t = o_t \odot tanh(c_t) \quad (8)$$

## C. Gated Recurrent Unit (GRU)

GRU, proposed by [19] is arguably a more simplified version of LSTM due to fewer gates, the GRU takes a linear sum between the existing state and the newly computed state similar to the LSTM but without having a separate memory cell. It operates on the so-called Update and Reset gates. The Sigma notation above represent those gates: which allows a GRU to carry forward information over many time periods in order to influence a future time period.

The GRU, however, does not have any mechanism to control the degree to which its state is exposed, but exposes the whole state each time. The GRU output unit is modulated by tanh like LSTM but lack $\tilde{c}_t$ for controlling hidden states.

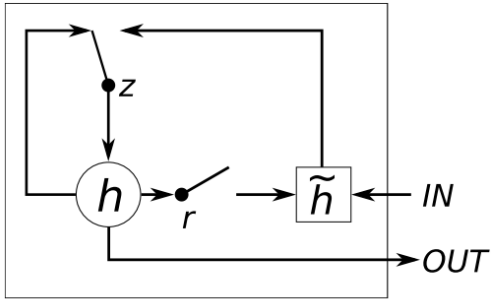

Fig. 3. Gated Recurrent Unit (GRU), where h is activation and h' is the candidate activation, while r and z are reset and update gates respectively

The equations of GRU units at each time step *t* takes input from previous memory cell $i_{t-1} = h_{t-1}$ and concatenate with the input $i_t$, the output is modulated with tanh to compute the potential candidate for hidden state $\tilde{h}_t$. The $r_t$ compute the relevance of $h_t$ for $\tilde{h}_t$. The potential candidate depends on $z_t$ which is relatively close to [1 or 0] from sigmoid activation, if its 1, $h_t$ gets updated else, no update and it maintains the previous content of $z_t$.

$$z_t = \sigma(W_z x_t + U_z h_{t-1}) \quad (9)$$
$$r_t = \sigma(W_r x_t + U_r h_{t-1}) \quad (10)$$
$$\tilde{h}_t = tanh(W x_t + U[r_t \odot h_{t-1}]) \quad (11)$$
$$h_t = (1 - z)h_{t-1} + z_t \tilde{h}_t) \quad (12)$$
$$i_t = h_t \quad (13)$$

The $\tilde{h}_t, z_t, h_t$ have the same dimensions, the $\odot$ element wise multiplication give the final output to activation function such as relu, softmax or sigmoid depending on the defined task.

Despite the fact that LSTM and GRU are relative similar, [16] were unable to conclude which gating units would perform better in general. LSTM have more gates which give more control during training but can slow down and consume more resources. While GRU is faster with a unit less than LSTM but without internal memory state for longer sentences[20]. This study is a step further in investigating the performance of LSTM and GRU on various types of text classification tasks.

## III. TEXT CLASSIFICATION TASKS AND ARCHITECTURE

Single labelled text classification task assigns a class to a document according to their content from a previously defined taxonomy (a hierarchy of classes) [7] [21]. More formally, if document *d* can belong to a fixed set of classes $C = \{c_1, c_2, \ldots, c_n\}$ then a training set of *m* labelled documents $\{d_1, c_1\}, \ldots, (d_m, c_m)$ output a learned classifier $\hat{y}: d \rightarrow c$.

Text Classification is broadly approached in two ways: binary and multiclass.

### A. Binary Classification

This type of classification is also known as binomial classification, is a type of classification task that outputs only one out of two mutually exclusive classes. For example, in medical testing, a learning model to detect HIV presence and outputs either "positive" or "negative". It is popularly used for email spam detection[22], [23] and sentiment analysis [24]. During training, we used sigmoid classifier at the last layer of the deep neural network.

### B. Multiclass Classification

This classification type is also called multinominal classification, this task distinguishes among more than two predefined categories or classes. For example, categorizing Wikipedia articles into (arts, history, law, medicine, religion, sports and technology) [6]. However, multi-class problem can also be transformed into binary class problems through decomposition. A close variation of this type of classification is Multi-label [25], where a particular input can belong to more than one class such as Visual Dialog.

### C. Architectural Framework

In the proposed framework, the inputs are usually unstructured texts with labels to be trained. These inputs are of varying lengths and with different English words and vocabulary composition. English language models are investigated to have prior knowledge of the overhead cost in order to balance the accuracy, storage, memory with the task's specification.

Data preprocessing remove stopwords, clean unwanted characters and symbols that add less meaning to the overall sentences, and capitalization where necessary. The inputs before training need to be uniform before feeding into the neural network via mean deviation of the variable length sentences. Truncation of sentences that exceed the threshold and padding of short sentences with <UNK>.

…

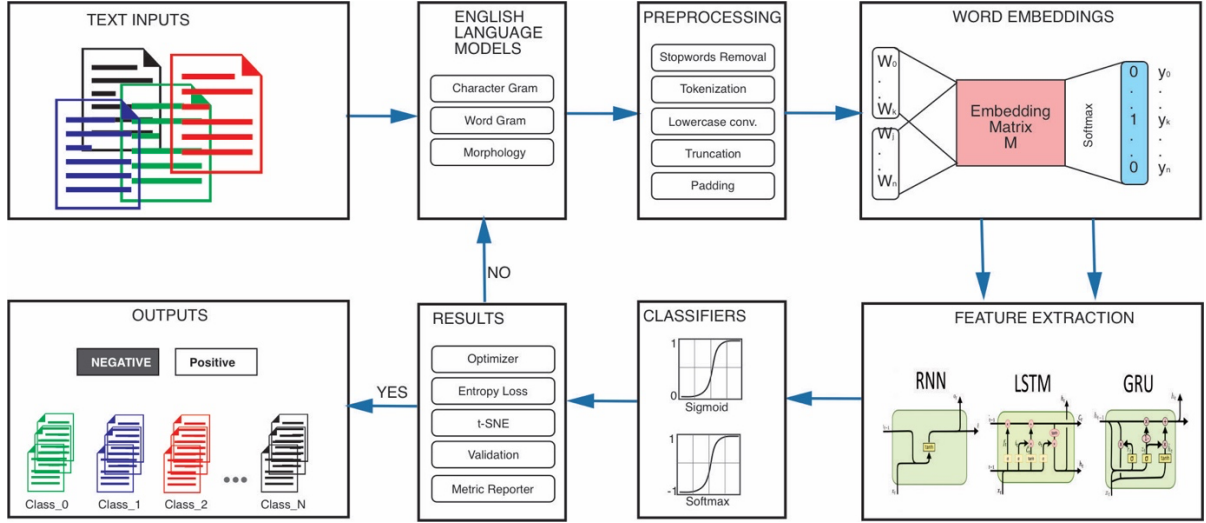

Fig. 4. Overall Architectural Framework with Various Stages of the Model

The outputs from the Preprocessing stage are tokenize into words. These words are in turn represented with vectors using featurized representation and one-hot encoding multiplied with embedding matrix to form word embeddings. It assigns a high multi-dimensional value to represent each word of the inputs. In the case of this work, we used 300D Dimensional Vectors.

The Feature Extraction stage is the core of the framework where the neural network learn to find patterns in the texts by iteratively adjusting the weights and biases through backpropagation. RNN, GRU and LSTM are the candidates used for these tasks of pattern/feature extraction.

To arrive at an acceptable model, we iteratively fine-tune the model by adjusting the hyperparameters based on loss and optimization functions specified. Optimizations used include Adam, RMSProp and Mean Square Error. The loss functions are Binary Crossentropy, Categorical Crossentropy and Sparse Categorical Crossentropy. In every single training, the outputs are classified with Sigmoid (binary) or Softmax (multi-class) and the accuracy of the model is calculated.

## IV. TRAINING

For binary classification, the first layer is a shallow network of pre-trained embeddings with the dimension (Vocab_size by Embedding_Dim i.e. 10000 by 16), the hidden layers consist of 16 and 8 neurons respectively. The ultimate output responsible for classification is fed into dense layer with one neuron and sigmoid activation.

$$\hat{y} = \frac{1}{1 + e^{-(W^{[L]}a^{[L-1]} + b^{[L]})}} \quad (14)$$

We compare the actual output ($\hat{y}$) with the label output (y) to compute the loss by estimating the maximum likelihood

$$\mathcal{L}(\hat{y}, y) = -y \, log(a) - (1 - y) \, log(1 - a) \quad (15)$$

The global cost function minimized to the barest with

$$\mathcal{J}(W, b) = \frac{1}{m} \sum_{i=1}^{m} \mathcal{L}(\hat{y}^i, y^i) \quad (16)$$

Optimizing with Gradient Descent to find the global minima [26], we backpropagate the cost function to update the new Weights and bias at each iteration:

$$W^{new} = W^{old} + \frac{\partial \mathcal{L}}{\partial a} \frac{\partial a}{\partial z} x \quad (17)$$

$$b^{new} = b^{old} + \frac{\partial \mathcal{L}}{\partial a} \frac{\partial a}{\partial z} \quad (18)$$

The multiclass classification used number of classes (C) with softmax activation function. The output is computed with the following equation:

$$\hat{y} = \frac{e^{(W^{[L]}a^{[L-1]} + b^{[L]})(i)}}{\sum_{1}^{C} e^{(W^{[L]}a^{[L-1]} + b^{[L]})(C)}} \quad (19)$$

Maximum likelihood estimation of softmax in the classification involving more than two classes requires summation of individual candidate class, the winner class with 1 end up being summed up and the others equal to 0.

$$\mathcal{L}(\hat{y}, y) = -\sum_{j=1}^{C} y_j \, log \, \hat{y}_j \quad (20)$$

## V. EXPERIMENTS

### A. Dataset Preparation and Preprocessing

To demonstrate the effectiveness of sequence models, each classification type has a specific dataset. The datasets are briefly described as follows:

- IMDB: The IMDB dataset is a movie reviews labeled into two binary classes [Positive and Negative]. We train our model on a subset of 50,000 reviews, each containing one or more sentences. The dataset is split equally for training and testing.

- BBC News Corpus: This dataset consists of news articles belonging to one of these five classes [Sport, Business, Politics, Entertainment, and Health]. Summary of Dataset after preprocessing

| Dataset | Type | Train size | Test size | Class | Averaged length | Vocabulary size |
|---|---|---|---|---|---|---|
| IMDB | Document | 25,000 | 25,000 | 2 | 250 | 385K |
| BBC | Document | 4,500 | 3,000 | 5 | 400 | 100K |

As typical text models rely on a vector representation for each word, we used a fixed vocabulary for all the tasks. Every out-of-vocabulary word was replaced with a special "EOS" token. We padded the short sentences with pre-zeros and truncate sentences that are too long at the end.

### B. Hyperparameter Tuning

The classification models are trained with backpropagation [27] and word2vec [28] for generating the word embeddings. We took the fourth root of the total vocabulary size to get an estimated number of 16 for the embedding dimension. The categorical numerical values to our inputs get assigned automatically during training after transferring learnt knowledge from previous trained word vectors.

The Weights and Biases are initialized randomly with numbers slightly above 0 (between 0.0 – 1.0). The hidden layers consist of the sequence models, including LSTM, RNN and GRU. The initial learning rate is 0.001 binary classifier network and 0.005 for multi-class classifier network.

### C. Training

Experimentally, The learning curves of the trained network are displayed in Fig. 5 and Fig. 6. For Exact value, TABLE I. list all the results from our experiments.

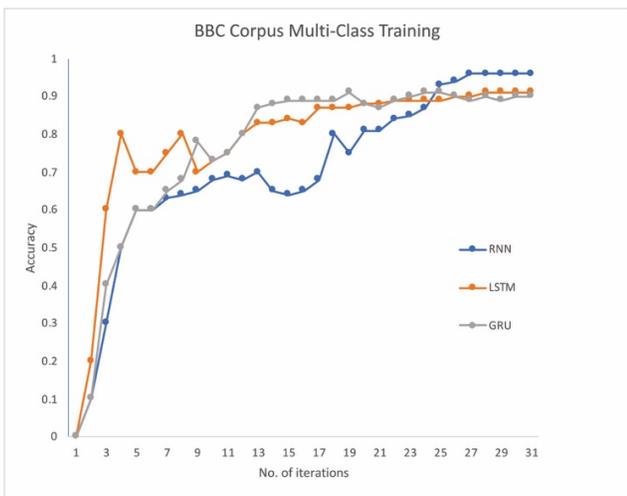

Fig. 5. Learning Curves during training for (BBC) Multi-class Classifier Network on Sequence models.

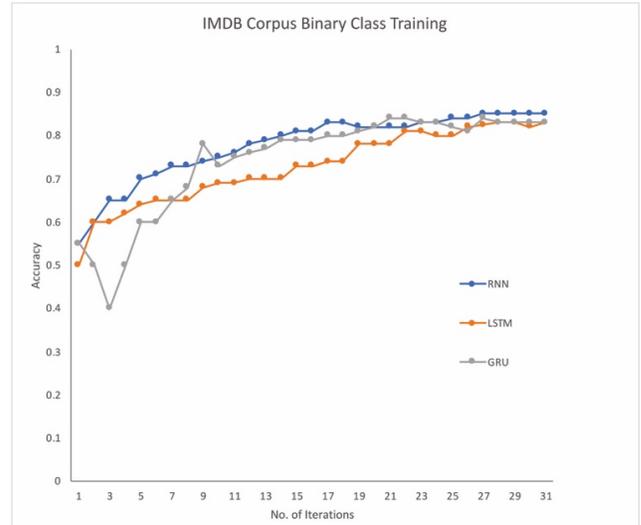

Fig. 6. Learning Curves during training for (IMDB) Binary Classifier Network on Sequence models.

Generally multi-class models on BBC converged better and faster with optimum accuracy of 96% and loss of 2%. Binary tasks achieved 88% accuracy and loss of 4% after 30 epochs. All the sequence models performed competitively with excellent accuracy and minimal loss.

## VI. EVALUATION AND ANALYSIS

### A. Evaluation

To determine the effectiveness of the various classification tasks, we used precision, recall, and the harmonic mean F1 score [30]. These scores are used to find classes that get confused with others to develop features to distinct them from others.

TABLE I. PERFORMANCE METRICS ON IMDB DATASET WITH BINARY CLASS

| Model | Accuracy | Precision | Recall | F1 Score |
|---|---|---|---|---|
| RNN | 86.42 | 88.30 | 85.10 | 86.67 |
| GRU | 85.54 | 87.51 | 84.19 | 85.82 |
| LSTM | 85.16 | 86.27 | 84.40 | 85.32 |

TABLE II. PERFORMANCE METRICS ON BBC DATASET WITH MULTI-CLASS

| Model | Accuracy | Precision | Recall | F1 Score |
|---|---|---|---|---|
| RNN | 96.18 | 96.28 | 96.18 | 96.19 |
| GRU | 90.34 | 91.00 | 90.34 | 90.16 |
| LSTM | 91.46 | 91.49 | 91.46 | 91.44 |

The results from this study portray an outstanding implication. From the performance metrics it is evident that the sequence models are performing excellently in both binary and multi-class classification tasks. Amazingly, performance on BBC multi-class was even better with more than 90% accuracy in most cases. More specifically, RNN achieved F1 of 96.19% on BBC multi-class task while on IMDB binary task, the F1 score is slightly lower with 86.67%.

Performance of RNN surpass GRU and LSTM both in binary and multi-class tasks. Furthermore, GRU performed better than LSTM in binary task on IMDB dataset; while LSTM out-performed GRU in multi-class tasks on BBC dataset. Overall, we were surprised that sequence models are doing excellently in classifying inputs consisting of long sentences.

## VII. Conclusion

This research covered several aspects of deep learning in natural language processing. We explored in depth the various sequence models in deep learning that can be used for text classification including RNN, LSTM and GRU. One thing these models have in common is the ability to recall previous information to exploit long range dependencies.

We analyzed in details Binary and Multiclass text classification with respect to single labelled data. Binary classification task was trained on IMDB dataset with sigmoid classifier, while Multi class classification was trained on BBC news dataset with softmax classifier. In order to accurately classify the documents, both classification tasks were trained on deep sequence networks.

Different models have different parameter configuration which can influence the output of the network such as learning rate, epoch, optimization, regularization, dropout, size of hidden layer. After some tuning, a very interesting result we obtained based on F1 and Accuracy showed that RNN are still relevant when it outperformed both LSTM and GRU in all the classification tasks. In addition, GRU converged better on IMDB Binary classification task while LSTM was better on Multi-class classification tasks. However, these hyperparameter tuning is not exhaustive as we strongly believe it can be fine-tuned to get improved results in the future.